\pdfoutput=1
\documentclass[11pt]{article}

\usepackage{EMNLP2022}

\usepackage{times}
\usepackage{latexsym}
\usepackage{graphicx}

\usepackage[T1]{fontenc}
\usepackage{booktabs}

\usepackage[utf8]{inputenc}
\usepackage{hyperref}
\usepackage{microtype}
\usepackage{multirow}

\usepackage{tablefootnote}
\usepackage{inconsolata}
\title{
Borrowing Human Senses: Comment-Aware Self-Training for \\Social Media Multimodal Classification
}

\author{Chunpu Xu, Jing Li\thanks{~~~Corresponding author}
\\ % All authors must be in the same font size and format. Use \Large and \textbf to achieve this result when breaking a line
Department of Computing, The Hong Kong Polytechnic University, China\\
\texttt{chun-pu.xu@connect.polyu.hk}\\ \texttt{jing-amelia.li@polyu.edu.hk}}

\begin{document}
\maketitle
\begin{abstract}
%Millions of cross-media posts consisting of image-text pairs are created on social platform every day to convey users' ideas and feelings.
%\jing{Social media is daily exhibiting huge volume of cross-media posts in the form of  image-text pairs.}
% Thus, amounts of multimodal classification tasks are derived to analyse social media for different applications.
% \jing{Social media is daily exhibiting massive multimedia content with pair-wised image and text.
% It hence presents the pressing need to automate the vision and language understanding with various multimodal classification tasks.}
Social media is daily creating massive multimedia content with 
paired image and text, presenting the pressing need to automate the vision and language understanding for various multimodal classification tasks.
%The advance of multimodal classification would crucially automate the vision and language understanding 
%on social media exhibiting massive multimedia content every day.
% However, different from traditional  multimodal classification tasks (i.g., visual entailment and visual question answering) where strong semantic interrelation are in the image-text pair, weak semantic correlation are contained in the cross-media posts. 
% Most existing methods focus on exploring the cross-modal interactions, despite of their implicit and intricate nature on social media.
% As a result, it's hard for the multimodal model to learn the interactions between images and texts since the content of images and texts are not aligned.  
% To tackle the problem, we use the comments obtained from similar posts as the bridge to connect the image modality and text modality.
Compared to the commonly researched visual-lingual data, social media posts tend to exhibit more implicit image-text relations.
To better glue the cross-modal semantics therein, 
% \jing{which tend to be implicit on social media}, 
%which tend to be implicit on social media, 
we capture hinting features from user comments, which are retrieved via jointly leveraging visual and lingual similarity.
% Additionally, the size of most datasets of multimodal social media tasks are small due to the expensive annotation cost. To alleviate the problem, we present a self-training method, which employs pseudo-labeled data constructed from the  retrieved similar image-text pairs to improve learning. 
Afterwards, the classification tasks are explored via self-training in a teacher-student framework,
%based on pseudo-labeling, 
motivated by the usually limited labeled data scales in existing benchmarks.  
%of labeled multimodal data for social media tasks. 
%multimodal classification. 
%labeled data in most multimodal social media datasets.
%We evaluate our approach 
Substantial experiments are conducted on four multimodal social media benchmarks for image-text relation classification, 
%multimodal 
sarcasm detection, %multimodal 
sentiment classification, and 
%multimodal 
hate speech detection.
%In the experiments, 
The results show that our method further advances the performance of
%enables better results of 
previous state-of-the-art models, which do not employ comment modeling or self-training.
\end{abstract}

\section{Introduction}

%\paragraph{introduction to multimodal social media}
% The growing popularity of multimedia is revolutionizing the  communications on social media.
Interpersonal communications in multimedia are gaining growing popularity on social media.
% The conventional text-only form has been expanded to cross modalities involving texts and images in information exchange.
%Many online platforms, such as Twitter, allow easy posting with both images and text.
More and more social media users are turning to pair images to text and vice versa to better voice opinions, exchange information, and share ideas, exhibiting rich and ever-updating resources in multimedia.
% Multimodal social platforms such as Twitter make users easy to create content by integrating texts and images.
%It is substantially because many online platforms, such as Twitter, allow easy posting with both images and text.
% \jing{Here is a gap: we should further discuss why we need to automate multimedia analysis, e.g., better use of the social media resources compared to the massive amounts of multimedia posts.}
%Millions of users together create 
While potentially benefiting people's everyday decision making, the huge volume of multimedia content might also challenge users in finding what they need.
Towards a more efficient and effective way to process the online multimodal data,
%enable easy access of the information in need for individuals, 
substantial efforts have been made to automatically understand the vision and language on social media through 
a broad range of  
multimodal classification tasks for predicting
%, e.g.,
%For better use of the social media resources compared to the massive amounts of multimedia posts,
%analyzing the multimodal posts is valuable for the areas of computional social science and derives many different multimodal classification tasks (i.g., 
image-text relations 
%classification 
\cite{DBLP:conf/acl/VempalaP19}, 
%multimodal 
sarcasm 
%detection 
\cite{DBLP:conf/acl/sarcasm}, 
%multimodal 
metaphor 
%understanding 
\cite{DBLP:conf/acl/ZhangZZ0L20}, point-of-interest 
%prediction
\cite{DBLP:conf/emnlp/VillegasA21}, 
%multimodal 
hate speech 
%detection 
\cite{DBLP:conf/acl/BotelhoHV21}, 
%and 
%multimodal 
sentiment 
%classification 
\cite{DBLP:conf/ijcai/Yu019}, etc.
%in social media.

Despite the success of visual-lingual understanding witnessed in common domains \cite{DBLP:conf/acl/HuangWCZCLL20, shi-etal-2020-improving, DBLP:conf/emnlp/WangJSYS21}, existing models' performance is likely to be compromised on social media posts. 
The possible reason lies in the relatively more implicit and obscure image-text relations therein \cite{DBLP:conf/acl/VempalaP19}, whereas the image-text pairs in the widely-used datasets outside social media (e.g., COCO dataset \cite{DBLP:conf/eccv/COCO}, VQA dataset  \cite{DBLP:conf/iccv/VQA}, VCR dataset \cite{DBLP:conf/cvpr/VCR}) tend to present explicit information overlap.
Such issue is nevertheless ignored in many previous solutions, which follow the common practice to fuse visual and lingual features \cite{DBLP:conf/acl/VempalaP19,DBLP:conf/acl/ZhangZZ0L20,DBLP:conf/emnlp/HesselL20,DBLP:conf/acl/BotelhoHV21}, making it hard for a multimodal model to well align cross-modal semantics attributed to their weak correlations \cite{nature2022_weak_semantic}. 

%\paragraph{The weakness of previous work (ignore the gap in img-text pair)}
%As indicated by \cite{DBLP:conf/acl/VempalaP19}, images and texts usually play different roles when expressing the overall multimodal posts, and the semantic relationships are split into four types based on the information overlap, which means there are often significant gaps of content in the image-text pairs. 
%However, most previous work (i.g., \cite{DBLP:conf/acl/VempalaP19}, \cite{DBLP:conf/acl/ZhangZZ0L20}, \cite{DBLP:conf/acl/BotelhoHV21}) ignore this potential problem and directly fuse the image features ans text features to understand the overall meaning of posts. 
%It's hard for the multimodal model to learn the alignment between the images and texts due to the weak semantic correlation \cite{DBLP:journals/corr/wenlan}. 

 \begin{figure}[t]
\centering
\includegraphics[width=1.0\columnwidth]{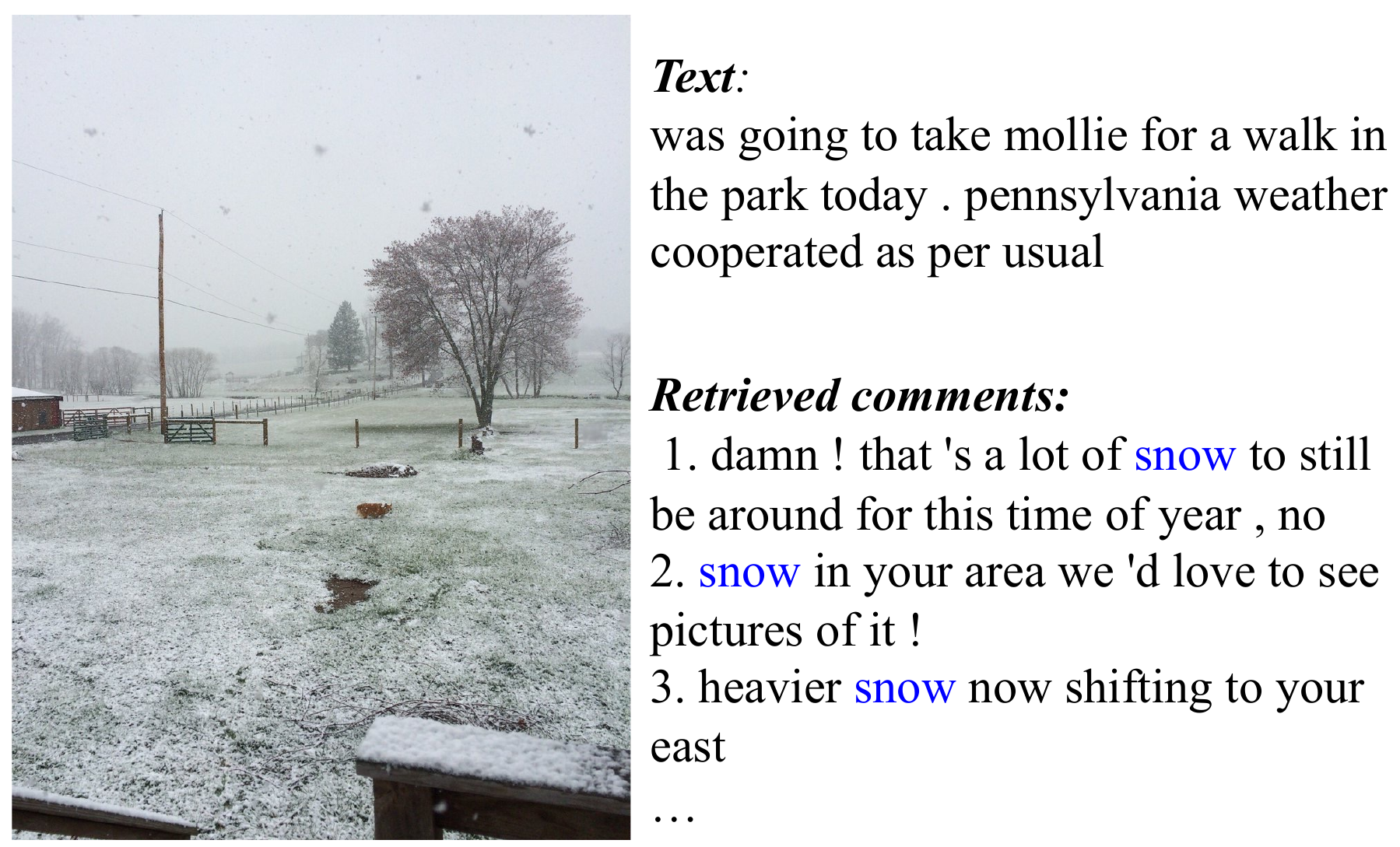}
\vspace{-2em}
\caption{
A sample tweet with its image on the left. On the right,  the tweet text is shown on the top, followed by the comments retrieved from similar tweets.
The word ``\textcolor{blue}{snow}''  (in blue) in comments helpfully hint the implicitly shared semantics between image and text.
% \jing{We may boldface the titles, i.e., ``Text'' and ``Retreved Comments''. And color the keyword ``snow'' in comments with blue.} \xu{replace the case}
%The proposed retrieval-based comment-aware self-training framework.
%representations encoded from texts (bottom), captions (upper left), and images (upper right).
%The encoded captions and texts are compared at output layer in visual-textual contexts.
%Multi-head attentions are employed to explore text-caption and image-text interactions. 
%The attended images, captions, and max-pooled texts are integrated to predict the discourse labels.
}
\vspace{-1em}
\label{fig:intro}
\end{figure}

Nonetheless, human readers seem to have no problem in digesting the cross-modal meanings on social media;
%, regardless of the possibly implicit image-text relations;
in response to what they capture from a post, some readers may drop user comments, where some clues to hint cross-modal understanding may be hidden, e.g., an echo of the keypoints.
For instance, in Figure \ref{fig:intro}, ``snow'' in retrieved comments might strengthen the connection of ``weather'' in text with the snowing visual scene.
%\jing{We may discuss Figure 1 here.}
The helpfulness of user comments has also been demonstrated in the previous NLP practice for text-only posts \cite{DBLP:conf/naacl/WangLKLS19}.
Inspired by that, we propose ``borrowing'' the senses from human readers and modeling user comments to learn the hinting features therein to bridge the image-text gap.
%\paragraph{Retrieve comments to bridge the gap between img and text}
%To strengthen the semantic correlation, we utilize the comments posted by other users to fill the content gaps. 
%The comments under original posts usually contain the understanding of human readers to the posts, and could provide a bystander perspective to highlight the important content and bridge the potential correlation in the multimodal posts. 
%Due to that comments are missing in most multimodal social media datasets and not all multimodal posts contain comments, 
To further benefit posts without user comments, a comment retrieval algorithm is designed to gather comments from other posts in similarity, which is measured via balancing the visual and lingual semantics (henceforth \textbf{cross-modal similarity}). 
For the related experimental studies,  a large-scale dataset  is constructed to mimic the open environment (henceforth \textbf{wild dataset}). It contains over 27M multimodal tweets, each with 3 comments on average.

% \jing{It seems that the statistics in intro may distract the readers when they just start understanding our stories. Probably we can put it in the dataset part.}
 
Then, we explore how to leverage the retrieved comments in multimodal classification and exploit a self-training framework to identify comments' hints which shape the cross-modal understanding (henceforth \textbf{comment-aware self-training}).
This considers method feasibility in scenarios where large-scale labeled data is unavailable, which commonly appears in the realistic practice, because 
the annotation for multimodal data from social media is extremely expensive \cite{DBLP:conf/mm/MaYL0C19}.
Concretely, we adopt a teacher-student prototype \cite{DBLP:conf/emnlp/MengZHXJZH20, DBLP:conf/naacl/ShenQMSRH21} and tailor-make it to learn multimodal understanding with the help of user comments.
%\paragraph{Using self-training to generate pseudo-labeled data}
%Furthermore, large amount of supervised data are needed for multimodal classfication tasks to learn the interactions between image modality and text modality. However, manually annotating the labels for multimodal data is costly. Significant amount of multimodal unlabeled data could be obtained on the social platforms.
%Inspired by \cite{DBLP:conf/emnlp/MengZHXJZH20, DBLP:conf/naacl/ShenQMSRH21} which use self-training to solve the task of text classification, we propose a retrieval-based multimodal self-training method to automatically annotate the unlabeled data in this paper. 
%Clearly, we first train
%a teacher model with the labeled data, and generate pseudo labels for the unlabeled data 
A teacher model is first trained with the labeled data and pseudo-label the similar posts with comments retrieved from the wild dataset. 
%based on cross-modal similarity.
%which is retrieved from the constructed large-scale multimodal social media dataset based on the similarity to the labeled data. 
Then, a student model is trained in guidance of both the knowledge gained by the teacher model and hinting features offered by user comments. %\jing{Comments were not mentioned in the original version. Check if the description is correct.}
%the labeled data and pseudo-labeled data are both utilized to train the student model. 

%\paragraph{summarize key observations in experiments}
%For empirical studies, 
To evaluate our method in practice, it is comprehensively experimented on four popular social media multimodal benchmarks for varying  classification tasks.
In the setup of each benchmark, our comment-aware self-training module is customized to BERT-based state-of-the-art (SOTA) architectures.
% , where the all SOTA performance have been advanced.
Ablation studies then exhibit the individual benefits provided by comments and self-training.
%that comments enables significant advance of all SOTA results forwards and self-training further boosts the model performance.
%datasets. 
%The proposed unified comment retrieval and self-training framework could be easy added to the state-of-the-art models. 
%The main comparison results on all datasets indicate that the model with retrieved comments could obtain significant improvement compared with the SOTA models without comments. 
%And utilizing the self-training architecture can further boost the performance of classification. 
Then, we analyze the effects of retrieved post number and find the use of more posts would result in both the benefits and noise.
%the num of retrieved unlabeled  image-text pairs to the performance. 
Next, we probe into comment retrieval and explore the contributions of visual and lingual modality in cross-modal similarity measure, where we observe the joint effects allow the best results.
%impacts of image modality and text modality to the multimodal retrieval process and show that  the modality fusion could achieve the best results. 
At last, a case study shows how the retrieved comments mitigate cross-modal semantic gap, followed by the an error analysis to discuss the existing limitation.
%a case study is demonstrated how the retrieved comment bridge the text modality and image modality.

In summary, our contributions are three fold.

$\bullet$ We demonstrate the potential to employ retrieved user comments from similar posts for a better visual-lingual understanding on social media and gather 27M cross-media tweets with comments to released to support future research in this line.
% \footnote{The code and dataset will be released upon publication to support the future work in the effects of user comments over the learning of social media multimodal understanding and beyond.}
\footnote{Our code and dataset are released at \url{https://github.com/cpaaax/Multimodal_CAST}.}

$\bullet$ A comment-aware self-training method is proposed for cross-modal learning from both human senses underlying retrieved comments and knowledge distilled from labeled data in limited scales.

$\bullet$ An empirical study with substantial results is provided, where SOTA models of four popular social media benchmarks for multimodal classification perform better with the help of our comment-aware self-training module and the retrieved comments bridge social media images and text via hinting the connecting points for models to attend to.

%propose a   comment-aware self-training framework for social media multimodal classification, where the retrieved comments are utilized to connect the semantic gap between image and text, and the retrieved image-text pairs are used to generate pseudo-labeled data for self-training.

% \paragraph{summarize the  contributions}
% Three main contributions in the paper are summarized as follows:

% \begin{itemize}
% \item We propose a general unified retrieval-based comment-aware self-training framework for social media multimodal classification, where the retrieved comments are utilized to connect the semantic gap between image and text, and the retrieved image-text pairs are used to generate pseudo-labeled data for self-training.

% \item A large-scale dataset consists of about 26 million cross-media tweets are collected, where each tweet contains at least one comments.

% \item Comprehensive experimental results on three social media multimodal datasets show that the proposed method could significantly improve the accuracy of classification and achieve the SOTA performance.
% \end{itemize}

\section{Related Work}

Our work is in line with multimodal learning and self-training, which are discussed below in turn.

%\subsection
\paragraph{Multimodal Learning.}
Previous work in this field focuses on
%Multimodal learning, 
%aiming at process and 
fusing features 
%extracted 
from different modalities (e.g., vision and language) \cite{DBLP:journals/pami/BaltrusaitisAM19} to tackle cross-modal 
%It has drawn growing 
%great 
%attention and has been widely used in conventional multimodal 
classification tasks, such as visual question answering (VQA) \cite{DBLP:conf/cvpr/TapaswiZSTUF16, DBLP:conf/cvpr/GoyalKSBP17, DBLP:conf/cvpr/JohnsonHMFZG17}, visual commonsense reasoning \cite{DBLP:conf/cvpr/VCR, DBLP:conf/nips/ViLBERT}, and image-text  retrieval \cite{DBLP:conf/eccv/LeeCHHH18, DBLP:conf/aaai/Unicoder-VL, DBLP:conf/eccv/Oscar}. 
%The image-text pairs in the dataset of these tasks 
Most benchmark data in vision and language assumes strong image-text correlations, and many multimodal models are hence designed to explore the common semantics shared by the two modalities.
%strongly correlated 
%and the multimodal models 
%are required to learn the semantic alignment contained in the cross-media data. 
However, it has been recently pointed out that many real-world scenarios, including social media, tend to present image-text pairs with weak and intricate cross-modal interactions  \cite{DBLP:conf/acl/VempalaP19,DBLP:conf/emnlp/HesselLM19,nature2022_weak_semantic}.
%as indicated by \cite{DBLP:journals/corr/wenlan}, the strong correlation assumption is invalid in real-world scenarios (i.g., social media) and the implicit interaction usually appears in the image-text pairs. 
%Additionally, \cite{DBLP:conf/acl/VempalaP19} 
%analyzes and defines the image-text relationships of social media posts into several types, which also illustrates existence of the weak correlation.

Despite the substantial efforts made in applying multimodal learning in social media to tackle various visual-lingual tasks, e.g., 
%In recent years, multimodal learning has also applied in social multimodal social media tasks, for instance, 
%multimodal 
sarcasm detection \cite{DBLP:conf/acl/sarcasm}, %multimodal 
hate speech detection \cite{DBLP:conf/acl/BotelhoHV21}, 
%mutimodal 
metaphor detection \cite{DBLP:conf/acl/ZhangZZ0L20}, etc., most existing methods follow the common practice to fuse visual and lingual features.
It is hence challenging for them to figure out the cross-modal meanings exhibit with implicit image-text links.
We thus resort to the retrieved user comments and study how models can find the hinting features therein to mitigate the cross-modal gaps.
%However, these work ignores the semantic gap in the cross-media posts, and it is challenging for the multimodal models to automatically learn the implicit correlation. 
%In the paper, we retrieve comments from similar posts to explicitly bridge the images and texts.

Our work is also related to \citet{DBLP:conf/emnlp/GurNSLKR21}, where the retrieved similar data show helpful in better aligning visual-lingual features for multi-modal classification. 
%where a cross-modal alignment module is trained for retrieving similar multimodal samples, 
%and the original input is augmented by concatenating the retrieved samples. 
Different from them, we additional retrieve comments and learn the hints therein via  self-training, which allows easy integration to most multi-modal classification archetectures.
%that, the  retrieval module designed in our work could adaptively retrieve similar posts and related comments by considering the weak semantics in the image-text pairs. 
%And the retrieved posts are involved in the self-training to improve the performance.

%\subsection
\paragraph{Self-training.}
Our method is inspired by previous work in
self-training
%Self-training
\cite{DBLP:journals/tit/Scudder65a, DBLP:conf/acl/Yarowsky95}, where labeled data is employed to train models and generate pseudo-labels for unlabeled data.
It is simple yet effective to enable model robustness with limited labeled data, potentially helpful in multimodal social media tasks owing to the expensive annotation and small-scale labeled data in most benchmarks \cite{DBLP:conf/acl/VempalaP19, DBLP:conf/acl/BotelhoHV21}.

Here we adopt the trendy self-training paradigm in a teacher-student framework, where a teacher model is trained with labeled data and self-label unlabeled data to generate synthetic data for the student model training.
It has been widely used in many tasks in CV (e.g., image classification \cite{DBLP:conf/cvpr/XieLHL20, DBLP:conf/nips/ZophGLCLC020}, object detection \cite{DBLP:conf/cvpr/YangWW0021}) and NLP (e.g, question answering \cite{DBLP:conf/naacl/SachanX18, DBLP:conf/emnlp/ZhangB19, DBLP:conf/emnlp/RennieMMNG20}, text classification \cite{DBLP:conf/nips/MukherjeeA20, DBLP:conf/emnlp/MengZHXJZH20, DBLP:conf/naacl/ShenQMSRH21}).
However, in existing work, limited attention has been drawn to its effectiveness in social media multimodal classification and how it works with retrieved comments to gain the cross-modal understanding for noisy data.
%which is a simple and old semi-supervised method, 
%typically employs the teacher-student framework, 
%where a teacher model trained with annotated data is utilized to generate synthetic data based on the large amount of unlabeled data, and then a student model is trained with the labeled data and the synthetic data.
%While self-training method  has been widely used in CV (i.g., image classification \cite{DBLP:conf/cvpr/XieLHL20, DBLP:conf/nips/ZophGLCLC020} and object detection \cite{DBLP:conf/cvpr/YangWW0021}) and NLP (i.g, question answering \cite{DBLP:conf/naacl/SachanX18, DBLP:conf/emnlp/ZhangB19, DBLP:conf/emnlp/RennieMMNG20} and text classification \cite{DBLP:conf/nips/MukherjeeA20, DBLP:conf/emnlp/MengZHXJZH20, DBLP:conf/naacl/ShenQMSRH21}), few work utilize the self-training method to solve multimodal classification tasks. Different from previous work where the highest confidence data predicted by the teacher model on the whole unlabeled dataset are used as the synthetic data, we directly retrieve similar multimodal data based on the similarity to each image-text pair in the training set as the source of synthetic data before training the student model to improve the efficiency.

\section{Comment-Aware Self-Training}
\begin{figure*}[t]
\centering
\includegraphics[width=1.85\columnwidth]{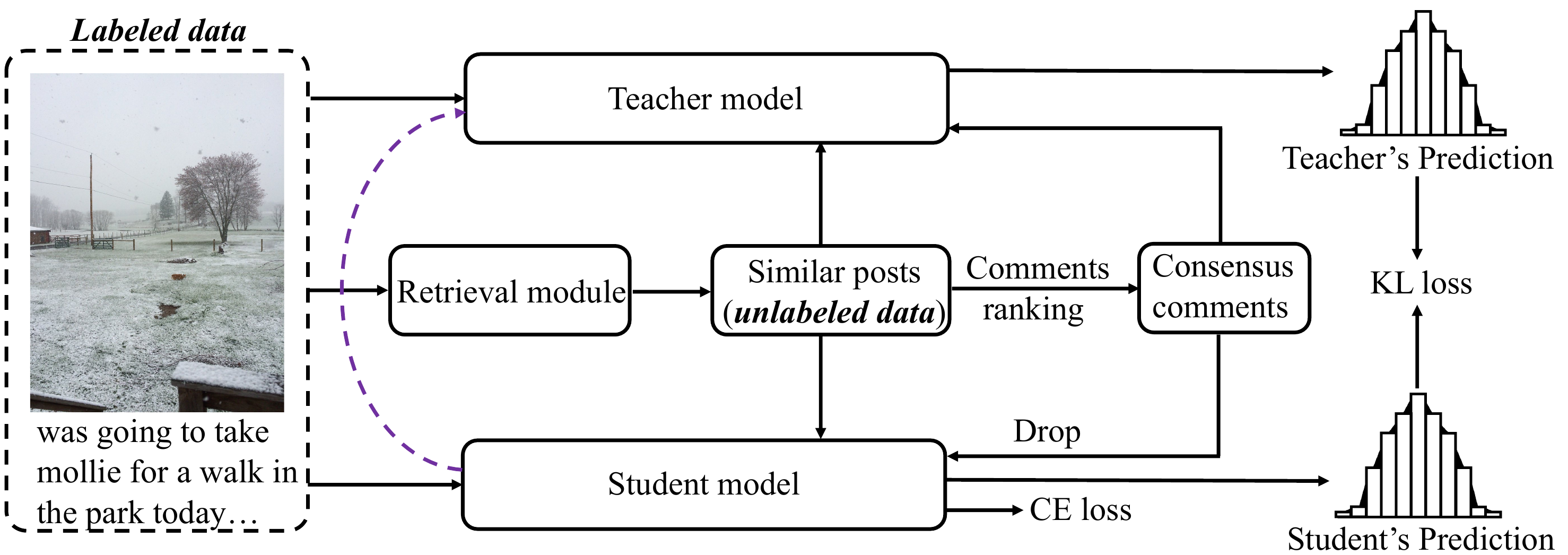}
\vspace{-0.5em}
\caption{
The workflow of comment-aware self-training. 
Given a post (image-text pair), we first query similar posts and their comments in a retrieval module.
Then the retrieved data is employed in teacher-student training as unlabeled data, where student model is trained with CE (cross-entropy)  and Kullback–Leibler (KL) divergence loss.
%The proposed retrieval-based comment-aware self-training framework. 
%The CE loss indicates the cross-entropy loss and KL loss represents the KL divergence loss. 
%representations encoded from texts (bottom), captions (upper left), and images (upper right).
%The encoded captions and texts are compared at output layer in visual-textual contexts.
%Multi-head attentions are employed to explore text-caption and image-text interactions. 
%The attended images, captions, and max-pooled texts are integrated to predict the discourse labels.
}
\vspace{-1em}
\label{fig:model}
\end{figure*}
This section presents the entire  comment-aware self-training workflow illustrated in Figure \ref{fig:model}.
In the following,
we first describe how we gather the wild dataset for retrieval, followed by the
%the details of the retrieval dataset 
%and conduct 
related analysis in $\S$\ref{ssec:method:dataset}. 
Then we introduce the retrieval module to 
%obtain 
find similar posts and their
%related 
comments in $\S$\ref{ssec:method:retrieval_module}. 
Next, we describe the usage of comments in the multimodal architecture in $\S$\ref{ssec:method:comment_usage}.
At last, $\S$\ref{ssec:method:self-training} presents the comment-aware 
%general 
%retrieval-based 
self-training framework based on retrieved results.

\begin{table}[t]
	 \centering
{\renewcommand{\arraystretch}{1.0}
\resizebox{1.0\columnwidth}{!}
{
	\begin{tabular}[b]{l cccc}
		\toprule
		\textbf{$\textbf{Year}$}        & \textbf{Num} & \textbf{Text len} & \textbf{Com len} & \textbf{Com num}  \\
		\midrule

	    2014 & 3,178,845  & 12.88   & 8.81   &  3.27  \\
	    2015 & 6,373,198  & 13.24    & 8.19   & 3.32 \\
	    2016 & 6,230,437  & 13.69    & 8.98   & 3.13  \\
	    2017 & 4,583,203  & 11.37    & 8.55   & 3.37  \\
	    2018 & 3,370,186  & 10.95    & 9.01   & 3.41  \\
	    2019 & 4,048,959  & 10.46    & 8.35   & 3.06  \\
	    \hline
		\textbf{$\textbf{Total}$} & 27,784,828 & 12.31 & 8.62 & 3.25 \\
		\bottomrule	
		\end{tabular}}}
	\vspace{-1.5em}
	\caption{Statistics of the wild dataset for retrieval. \textbf{Text len} and \textbf{Com num} indicate the average length (token number) in text and average comment number per tweet. \textbf{Com len} is the average length per comment. \label{tab:retrieval_dataset_analysis}
	} 
	\vspace{-1em}
\end{table}
\subsection{Wild Dataset Construction for Retrieval }\label{ssec:method:dataset}

To simulate retrieval from the open environment, large-scale visual-lingual tweets with comments are gathered to form a wild dataset.
The detailed steps are described in the following. 
First, 
%we followed \citet{DBLP:conf/emnlp/NguyenVN20} to 
we downloaded the large-scale corpus used by \citet{DBLP:conf/emnlp/NguyenVN20} to pre-train their BERTweet. %containing
%, we first 
%download the Twitter data grabbed from the 
The dataset contains general Twitter streams 
%from the Internet Archive library.\footnote{\url{https://archive.org}} 
and we removed non-English tweets with 
fastText library \cite{DBLP:conf/eacl/fastText}.
Then, text-only tweets were removed and for the remaining 60 million, 
%60
%without images to obtain cross-meida posts. 
%About 60 
%million tweets with images remained.
%are acquired through this process. 
%Next, we filtered out non-English tweets with 
%Similar to \cite{DBLP:conf/emnlp/NguyenVN20}, we use the 
%fastText library \cite{DBLP:conf/eacl/fastText}.
%to classify the language of the tweets and remove non-English tweets. 
%Afterwards, 
images and comments were gathered using 
%Next, we collect the images and comments related to the tweet using 
Twitter stream API.\footnote{\url{https://developer.twitter.com/en/docs/twitter-api}}
% based on the tweet link. 
Next, 
%The 
user mentions and url links 
%in tweet texts and the comments 
are converted to generic tags @USER and HTTPURL for privacy concern. 
Finally, tweets without comments were further removed, resulting in
%keep the cross-media posts which have at least one comment, and get the retrieval corpus of about 
27.8 million tweets with images, text, and comments.
% We construct a large-scale multimodal social media dataset where each post contains at least one comment and obtain comments for the query post from similar posts. 

The over-year statistics of wild
%retrieval 
dataset is shown in Table \ref{tab:retrieval_dataset_analysis}.
There exhibits an interesting observation --- 
%from Table \ref{tab:retrieval_dataset_analysis} 
%is that 
the average text length in recent years (2017-2019) is 
%recent years (i.e., 2017, 2018, and 2019) is 
obviously shorter than earlier.
%that of previous years (i.e., 2014, 2015, and 2016). 
It might be because images these years allow the provision of gradually richer information, partially taking the text's roles in multimedia communications.
%users gradually prefer to use images exchange inforamtion  
%are gaining habits to exchange information with images while multimedia communications are becoming more and more popular. 
%This might because users tend to utilize images to replace part of texts to express content.

\subsection{Post and Comment Retrieval}\label{ssec:method:retrieval_module}

Given a post in image-text pair, we then discuss how to retrieve similar posts and their comments.
%from the wild dataset.

\paragraph{Retrieval of Similar Posts}
Post similarity is measured via balancing the effects of the image and text features.
The former is learned with
%we employ
%each tweets in the wild dataset, 
%we adopt 
ResNet-152 \cite{DBLP:conf/cvpr/Resnet} pre-trained on the ImageNet \cite{DBLP:journals/ijcv/ImageNet} and we take
%to encode images in the wild dataset and 
%dataset as the image encoder and 
the output of final pooling layer for representation.
%to represent image features.
%of ResNet-152 
%is utlized to represent the image features. 
For text features, SimCSE is adopted because of its effectiveness in similarity measure \cite{DBLP:conf/emnlp/SimCSE}. 
%, which significantly improve the sentence embeddings on the text similarity tasks, is used to extract the text features of text representations. 
%Given a query image-text pair in the dataset of target task, ResNet-152 and SimCSE are also employed to encode the image and text, separately. 
% Suppose that $s_i^I$ reflects the similarity measure 
% Then the 
For any post (the query), we score its similarity to the $i$-th multimedia post in the wild dataset with $s_i$: 

%semantic similarity score $s_i$ between the query post and the $i$-th post in retrieval dataset is calculated as:

\begin{equation}\small\label{eq:post-retrieval}
s_{i} = \alpha s^{I}_{i} + (1-\alpha) s^{T}_{i}
\end{equation}

\noindent where $s^{I}_{i}$ and $s^{T}_{i}$ respectively indicates the image and text similarity, traded-off by the parameter $\alpha$.

%indicates the similarity between the query image and the $i$-th retrieval image, while $s^{T}_{i}$ represented the similarity based on text modality. $\alpha$ is the parameter to balance the effects of text image and text. 

Here the weight $\alpha$ (balancing image and text effects) is empirically estimated with the averaged statistics measured over the wild datasets (for retrieval) and all experimental data (as a query set).
%To make $\alpha$ automatically determined by the internal characteristics of image-text pairs  and adaptive for different multimodal tasks, $\alpha$ is computed by:

\begin{equation}\small
\alpha = \frac{T_{mean}}{I_{mean}+T_{mean}}
\end{equation}
\begin{equation}\small
I_{mean} = \frac{1}{MK}\sum_{m=1}^{M}\sum_{k=1}^{K} p_{m,k}
\end{equation}
\begin{equation}\small
T_{mean} = \frac{1}{MK}\sum_{m=1}^{M}\sum_{k=1}^{K} q_{m,k}
\end{equation}

\noindent where $p_{m,k}$ and $q_{m,k}$ respectively refer to the image and text similarity between the $m$-th query and its rank-$k$ most similar post retrieved with the corresponding modality. 
$M$ is the query set size and $K$ the cut-off number of retrieved posts to be selected. 
%is the image similarity score between the $m$-th query image and the most similar $n$-th retrieval image, while $q_{m,n}$ is used to measure the text similarity. $N$ indicates the selected num of the most similar post, $M$ represents the size of the query dataset. Thus, $I_{mean}$ and $T_{mean}$ represent the overall average similarity of the image and text, separately.

% Furthermore, we use the Faiss library \cite{johnson2019billion} for fast similarity search. Clearly, the Inverted File Index Product Quantization(IVFPQ) \cite{johnson2019billion} index is built for feature vector compression and efficient kNN queries. IVFPQ index is trained with the images features to perform clustering on the original feature vectors. Given a query image features, IVFPQ would return $L$-nearest\footnote{$L$ is set to 0.1M in the experiments.} image indices and related image similarity score. Similarly, $L$-nearest text indices and related text similarity score could be obtained. For efficient search, the post related to the overlap index between text indices and image indices would be regarded as candidate similar posts, and participated in searching the final similar posts by Eq.\ref{eq:post-retrieval}.

In this way, for any query, we rank posts in wild dataset by the 
%the multimodal posts of the overlap index by the semantic similarity score 
$s_i$ score (leveraging image and text similarity) and the top-$K$ most similar posts will be retrieved.
%acquire top-$K$ similar posts for each query post in the target dataset. 
%Specially, 

% Here we give the details of using the Faiss library \cite{johnson2019billion} for fast similarity search. 
Here the Faiss library \cite{johnson2019billion} is employed for fast similarity search.
Concretely, the Inverted File Index Product Quantization(IVFPQ) \cite{johnson2019billion} index is built for feature vector compression and efficient kNN queries. First, IVFPQ index is trained with the images features to perform clustering on the original feature vectors. Then, given a query image features, IVFPQ would return $R$-nearest\footnote{$R$ is set to 0.1M in the experiments.} image indexes and related image similarity score. Similarly, $R$-nearest text indexes and related text similarity score could be obtained. At last, to allow an efficient search, the post related to the overlap indexes between text indexes and image indexes would be regarded as candidate similar posts, and used for searching the final similar posts by Eq.\ref{eq:post-retrieval}.

% Here the Faiss library \cite{johnson2019billion} is employed for fast similarity search, whose implementation details are put in Appendix \ref{appendix:similarity_search}.

\paragraph{Retrieval of Comments}\label{ssec:method:comment_retrieval}
As discussed above, comments (written by human readers)  potentially help in hinting the visual-lingual relations.
%Comments written by other users might be helpful to the understanding of the cross-media posts because a bystander's point of view is provided to bridge the semantic gap between the images and text. 
However, while building the wild dataset, we observe only 46\% multi-modal tweets contain comments. 
For those posts where comments are absent or inaccessible, the comments of the retrieved posts may be useful as well, because intuitively, similar posts may result in similar comments.

%on the one hand, not all posts contain comments. Actually, there are at least one comment in about 46\% cross-media posts in our observation. On the other hand, the comments are not collected and missing in most multimodal social media datasets. 
%Thus, we retrieve comments from similar posts based on the assumption that semantically similar cross-media posts might obtain similar comments.

However, because of the noisy nature of social media data, comments may vary in their quality and effects in hinting  cross-modal learning.
We therefore need to shortlist high-quality comments.

%The comments of top-$K$ similar posts are collected together and construct the comment pool $P$, where the comments are noisy since the comments are written from different human readers. 

Concretely, from the comment pool $P$ gathering all retrieved posts' comments, we follow   \citet{DBLP:journals/corr/consensus} to extract representative comments as consensus comments.
%consensus comments are defined to filter out noisy comments and select the representative comments which have 
They tend to exhibit relatively higher semantic similarity to other comments in $P$ and can be ranked with:
%And the consensus score is measured as follows:

\begin{equation}\small
q_{i}=\frac{1}{|P|}\sum_{p' \in P } Sim\left ( p_{i},p' \right )
\label{Eq:consensus}
\end{equation}

\noindent where $q_i$ is the consensus score of the $i$-th comment $p_i\in P$. 
%in the pool $P$. 
$Sim\left ( p_{i},p' \right )$ indicates the $p_i$-$p'$ similarity (with SimCSE).
%represents the similarity score between 
%the comment 
%$p_i$ and $p'$. 
$|P|$ is $P$'s comment size.
%is the num of 
%comments 
%in the pool . 

In practice, for any post, we query and retrieve
the top-$N$ consensus comments to
%with the highest consensus scores are obtained as 
form a set $C$ later used to bridge the gap between image modality and text modality (next discussed in $\S$\ref{ssec:method:comment_usage}).
%to engage in self-training.
%to bridge the semantic gap between images and texts. 
%The comments, encoded by BERT, are then injected into the attention mechanism fusing visual and lingual features to help explore cross-modal interactions.

%\jing{Still not very clear.}

%through the attention mechanism to interact with the text features and image features.

\subsection{Leveraging Comments in a Multimodal Classification Architecture}
\label{ssec:method:comment_usage}
The previous discussions concern how to retrieve the similar posts and their (consensus) comments. In the following, we describe the implementation details of how  comments can be leveraged in different multimodal classification architectures.  
Based on the different schemes for fusing image features and text features \cite{DBLP:conf/emnlp/AlbertiLCR19}, most multimodal classifiers could be divided into early fusion and late fusion. In the early fusion sheme, image features are embedded with text tokens on the same level (e.g., MMBT).  In the late fusion sheme, image features and text features are encoded separately and interacted by the concatenation (e.g., BERT-CNN) or the attention mechanism (e.g., CoMN-BERT and MMSD-BERT). 

Here we give the details of using comments to bridge the gap between images and texts without changing the original architecture of base classifiers. The comments injection methods are slightly different for early fusion and late fusion schemes, where the details come in the following.

\paragraph{Early Fusion Scheme}
The image features are projected into token space and concated with the word token embeddings as the input of multimodal bitransformer in MMBT. Then the hidden state $h^f$ related to the [CLS] token is used as the representation of the fusion vector for the classification in MMBT. We adopt the same encoding strategy to fuse each comment and the image features, and related hidden states $\{h^c_1,...,h^c_{N} \}$ are obtained. After that, the attention mechanism is used to compute the attended vector $u$, which is concated with the original fusion vector $h^f$ for the classification:
\begin{equation} 
u = \sum_{n=1}^{N}\beta_{n}h^c_{n}
\label{Eq:attention}
\end{equation}
\begin{equation} 
\beta_{i} = \frac{{\rm{exp}} (z_n)}{ \sum_{n=1}^{N}{\rm{exp}} (z_n)}; \quad z_n=\sigma(h^f, h^c_n)
\end{equation}
where $\sigma$ is a feed-forward neural network.

\paragraph{Late Fusion Scheme}
Assume the image features extracted by ResNet, text features extracted by BERT, and the original fusion vector, which is the output of the base classifier before the softmax layer, are $h^v$, $h^t$, and $h^f$, separately. And the comments features encoded by the BERT are denoted as $\{h^c_1,...,h^c_N \}$. Similar to Eq.\ref{Eq:attention}, then attention mechanism is employed to fuse the image features and comment features and obtain the image attended vector $v$. Similarly, the text attended vector $t$ which is fused by text features and comment features, could be acquired. At last, we concate the image attended vector $v$, text attended vector $t$, and original fusion vector $h^f$ for the final classification.

\subsection{ Self-training with Retrieved Posts}\label{ssec:method:self-training}

Here we further describe how the retrieved posts (i.e., the retrieved image-text pairs) is explored in multimodal classification.
Its data is commonly formulated as
%Formally, for conventional multimodal social media tasks like image-text relation classification, 
a labeled parallel dataset $L=\{x_i, c_i, y_i\}_{i=1}^{l}$, where $x_i$ is an image-text pair, $c_i$ indicates the retrieved $N$ comments, and $y_i$ a label specified by the task.
%is provided. 

The labeled dataset $L$ is usually limited in scales \cite{DBLP:conf/mm/MaYL0C19}, posing the over-fitting concern.
Meanwhile the retrieved posts, similar to the data in $L$, could form an unlabeled set ($U=\{x^{'}_i, c_i\}_{i=1}^{Kl}$) to enrich  training data. Note that $x^{'}_i$, one retrieved image-text pair of $x_i$, shares the same consensus comments $c_i$ with $x_i$.
Then $L$ and $U$ may be integrated to allow more robust learning in a semi-supervised manner.

%Compared with the  unlabeled retrieval dataset constructed in \ref{ssec:method:dataset}, the size of $L$ is usually small in most tasks. Thus, we utilize the top-$K$ similar posts obtained in \ref{ssec:method:retrieval_module} for each query post in training set to expand the scale of the training set. 
%The unlabeled dataset $U$ would be $K$ times the size of labeled dataset $L$. 
%The labeled data and the related similar retrieved unlabeled data share the same consensus comments. 
% Similar to labeled data, we also retrieve comments for each post in the unlabeled data.

Here we adopt self-training based on the popular teacher-student framework \cite{DBLP:conf/cvpr/XieLHL20}.  
A teacher model (the classifier) is first trained on the labeled data $L$ to gain task-specific knowledge and pseudo-label the unlabeled data with soft labels as ``teaching samples''.
Then a student model, sharing the same architecture as the teacher, is trained with both the pseudo-labeled $U$ and labeled $L$.

In the training of both teacher and student, their modality fusion mechanism is fed with the comment features (described in $\S$\ref{ssec:method:self-training}).
% , embedded via a BERT encoder.\footnote{Due to space limitation, we refer readers to the Appendix \ref{appendix:comments} for more implementation details of how comments are injected into varying architectures of multimodal classifiers.} 
It enables the models to explore cross-modal interactions in aware of the comments.

%which later educate the student model via pseudo-labeling the unlabeled data $U$.

%The self-training framework is adopted to utilize the labeled data and unlabeled data to improve the performance of the multimodal classification. 
%Clearly, a teacher model is trained with the labeled data integrated with related comments. 
%Then the trained teacher model is used to generate the soft labels, which are combined with the parallel data $L$ to train a student model. 

In practice, the student training randomly drops 50\% comments while teacher employs the full comment set.
It enables the student model to learn from the noised data for a better generalization instead of simply mimicking teacher's behavior.
%Additionally, the comments of labeled data and unlabeled data are dropped at the ratio of 0.5 when used for training the student model while the comments of labeled data are complete for training the teacher model. 
%Thus, the student model is enforced to learn from the noised input data and classify the accurate labels as much as possible

The teacher model is trained with the cross-entropy loss for classification while KL divergence loss is additionally used for student training:

%The student model is optimized with the cross-entropy loss and the KL divergence loss:
\vspace{-1em}
\begin{equation} \small
\mathcal{L} = \frac{1}{|L|} \sum_{i\in L} y_{i}\mathrm{log}y_{i} +\frac{1}{|U|} \sum_{i\in U} \mathrm{KL}(t_i||s_i)
\end{equation}

\noindent where $|L|$ and $|U|$ indicate $L$'s and $U$'s dataset size. 
%the size of labeled dataset and unlabeled dataset, separately. 
$t_i$ is the soft label predicted by the teacher model while $s_i$ is the output of the student model.

\section{Experimental Setup}

\begin{table}[t]
	 \centering
{\renewcommand{\arraystretch}{0.9}
\resizebox{0.9\columnwidth}{!}
{
	\begin{tabular}[b]{l rrrr}
		\toprule
		\textbf{$\textbf{Dataset}$}        & \textbf{\#Train} & \textbf{\#Val} & \textbf{\#Test} & \textbf{\#All}  \\
		\midrule

	    MVSA & 3,611  & 451   & 451   & 4,511  \\
	    ITR & 3,575  & 447   & 449   & 4,471 \\
	    MSD & 19,816  & 2,410   & 2,409   & 24,635  \\
	    MHP & 3,998  & 500   & 502   & 5,000  \\
		\bottomrule	
		\end{tabular}}}
	\vspace{-0.5em}
	\caption{Statistics of the evaluation datasets. \label{tab:four_dataset_analysis}
	} 
	\vspace{-1em}

	%\label{tab:results}
\end{table}

\subsection{Evaluation Datasets}

Our evaluation is conducted on four Twitter classification benchmarks on  
%We demonstrate our experiments on four public multimodal social media datasets: 
multimodal sentiment classification (MVSA) \cite{DBLP:conf/mmm/MVSA}, image-text relation (ITR) \cite{DBLP:conf/acl/VempalaP19}, multimodal sarcasm detection (MSD) \cite{DBLP:conf/acl/sarcasm}, and multimodal hate speech detection (MHP) \cite{DBLP:conf/acl/BotelhoHV21}. 
% ITR contains four classes data  which is annotated based on whether text or image provide additional content outside the other modality. MHP focus on the online hate scenario and  the label of the  
%The data in all datasets are 
Each data instance is an image-text pair 
%collected from Twitter 
and it is annotated with
%, and consists of a image-text pair and 
a single class label. 
For MVSA, MHP and MSD, we adopt the same 
%train, validation and test 
dataset split 
%for training, validation, and text 
as their original papers 
%\cite{DBLP:conf/acl/BotelhoHV21} and \cite{DBLP:conf/acl/sarcasm} 
for fair comparisons. 
For ITR, 
%lacking such setup details, 
we randomly
split 80\%, 10\% and, 10\% for training, validation, and test instead of their 10-fold cross-validation setup for the concern of experimental efficiency.

The statistics of evaluation datasets are shown in Table \ref{tab:four_dataset_analysis}, where we observe the small-scale training data in MVSA, ITR, and MHP.
For MVSA, though relatively larger in scales, its automatic labeling under hashtag-based distant supervision, may result in noisy labels, which further require larger data scales for robustness. 
These imply the annotation difficulties and potential benefits from self-training.

\subsection{Implementation and Evaluation Details}

All experimental models 
%for the experiments 
are implements with PyTorch\footnote{\url{https://pytorch.org/}} and HuggingFace Transformers\footnote{\url{https://github.com/huggingface/transformers}}. 
%The maximum length is 50 for both 

Both text and comment are capped at 50 words for encoding. 
The batch size is set to 8, 8, 16, and 16 for ITR, MHP, MVSA, MSD. 
The learning rate is set to 1e-5 with a warm-up rate to 0.1. 
Classifiers are trained with an AdamW optimizer. 
The maximum of consensus comments ($N$) is set to 5.
We run the self-training for three iterations. At each iteration,
the teacher model is fine-tuned for 10 epochs on the labeled training data. 
The teacher model performing the best in validation is adopted to predict the pseudo-labels for the unlabeled retrieved data. 
The student model is then 
fine-tuned for 10 epochs. After that, the student model is used as teacher for the next iteration.
%with the labeled training dataset and pseudo-labeled data. 

For evaluation metrics, we follow the benchmark practice to use precision (pre), recall (rec), and F1-score (F1) for
%to evaluate the performance of models for 
ITR, MVSD, and MHP, 
%tasks 
and accuracy (acc) and F1 
%is used 
for MSVA task.

\subsection{Baselines and Comparisons}

To investigate our universal benefits over different classification tasks varying in SOTA methods, we integrate our comment-aware self-training module into the BERT-based SOTA architectures and examine the results following baselines and comparisons employed in the original paper for fair comparison.
%For fair comparisons with previous work, different baselines are adopted for the four tasks due to the characteristics of different tasks 
%\footnote{I.g., the attributes of images are usually adopted for MSD while the texts in the images are extracted by using OCR for MHP.}.

%\paragraph{Baselines for MVSA.}

\paragraph{MVSA Comparisons.} This benchmark presents baselines of
MultiSentiNet \cite{DBLP:conf/cikm/MultiSentiNet} (a deep semantic network with the visual clues guided attention),
%mechanism 
%for multimodal sentiment classification
CoMN \cite{DBLP:conf/sigir/Co-MN} 
%is 
(a co-memory network to learn image-text interactions),
%the interactions between image features and text features
%for multimodal sentiment analysis. 
MMMU-BA \cite{DBLP:conf/emnlp/MMMU-BA} 
%proposes 
(enriching context for cross-modal fusion), and
%a fusion method by utilizing the context from neighboring utterances to generate richer multi-modal representation)
Self-MM \cite{DBLP:conf/aaai/Self-MM} (joint training of uni-modal and multi-modal tasks to explore cross-modal consistency).
%jointly learns the uni-modal tasks and multimodal sentiment analysis to capture the the consistency and difference among different modalities.) 
CoMN-BERT is a SOTA architecture combining CoMN and the pre-trained BERT \cite{DBLP:conf/naacl/BERT}, which will later be combined with our module for comparison.

%which employs the CoMN architecture and uses BERT \cite{DBLP:conf/naacl/BERT} 
%to encode text is used as our base model for MVSA.

%\paragraph{Baselines for ITR} 
\paragraph{ITR Comparisons.}
In the original paper \cite{DBLP:conf/acl/VempalaP19}, LSTM-CNN performs the best via combining CNN-encoded  visual features \cite{DBLP:conf/cvpr/InceptionNet} and LSTM-encoded lingual features \cite{LSTM}.
It is compared with the baseline ablations CNN and LSTM using uni-modal features only.
To line up with the SOTA, we implement BERT-CNN to employ pre-trained BERT for text encoding instead of LSTM, which is likewise compared to a BERT classifier using lingual features only.
Based on BERT-CNN, we integrate in our module to examine its effectiveness over SOTA. 
%Following \citet{DBLP:conf/acl/VempalaP19}, LSTM \cite{LSTM} and CNN \cite{DBLP:conf/cvpr/InceptionNet} are adopted to encode text and image, respectively.
%are adopted as the text-based method and image-based method for image-text relation classification, separately. 
%LSTM-CNN \cite{DBLP:conf/acl/VempalaP19} combines the image features and text features to jointly learn the interaction of image-text pairs. 
%BERT, a pretrained model for text, is also taking as the text-modality approach. BERT-CNN is utilized as our base model for ITR.

%\paragraph{Baselines for MSD}
\paragraph{MSD Comparisons.} Here the MMSD baseline is introduced in the original paper \cite{DBLP:conf/acl/sarcasm}, which employs a hierarchical fusion model to explore visual and lingual features with optical characters.
%\footnote{
%Here visual features include 
%Image attributes refer to optical characters in images.
%are usually adopted for MSD while 
%the texts in the images are extracted by using OCR for MHP.
%}
%introduces a hierarchical fusion model to learn the joint representation of text features, image features and image attributes. 
We also compare with the following more advanced models on the benchmark: D\&R Net \cite{DBLP:conf/acl/D_R} (using decomposition and relation network to learn visual-lingual interactions),
%to capture the semantic interactions between image modality and text modality.
Res-BERT \cite{DBLP:conf/emnlp/Pan_sarcasm} (concatenating visual features from ResNet (cite) and lingual features from BERT),
%(concates the image features and text features for sarcasm classification) 
Att-BERT \cite{DBLP:conf/emnlp/Pan_sarcasm} (with attention mechanism to capture image-text semantic consistency), and
%attention mechanism to capture the inconsistency between images and texts),
CMGCN \cite{DBLP:conf/acl/CMGCN} (building a  graph to explore cross-modal interactions).
%constructs a cross-modal graph to utilize the inconsistent implications between different modalities. 
MMSD-BERT is based on MMSD with a pre-trained BERT to encode texts, where we will later architect with our proposed module.

%which uses the NMSD architecture and uses BERT to encode text is used as our base model for MSD.

\begin{table}[t]
	 \centering
{\renewcommand{\arraystretch}{0.8}
\resizebox{0.7\columnwidth}{!}
{
	\begin{tabular}[b]{
	%l c c c c c c
	lcc
	}
		\toprule
		\textbf{Methods}        & \textbf{Acc} & \textbf{F1} \\
		\midrule

		MultiSentiNet
		& 69.84   & 69.63 \\
		%GRN (
        CoMN
        & 70.51 & 70.01 \\
        MMMU-BA
        & 68.72 & 68.35 \\
        Self-MM
        & 72.37 & 71.96 \\
        CoMN-BERT
        & 71.33 & 70.66 \\
        \midrule
        
        CoMN-BERT (full)
        &  \textbf{73.71}  & \textbf{72.83} \\
		\bottomrule	\end{tabular}}}
	\vspace{-0.5em}
	\caption{
	Comparison results 
	%of the baselines and our model 
	on the MVSA dataset. 
	}
	\label{tab:MVSA_results}
\end{table}
\begin{table}[t]
	 \centering
{\renewcommand{\arraystretch}{0.8}
\resizebox{0.8\columnwidth}{!}
{
	\begin{tabular}[b]{
	%l c c c c c c
	lccc
	}
		\toprule
		\textbf{Methods}      & \textbf{Pre}  & \textbf{Rec} & \textbf{F1} \\
		\midrule

		LSTM
		&42.33  & 48.55  & 38.77 \\
		%GRN (
        CNN
        & 37.11 & 47.22 & 35.99 \\
        LSTM-CNN
        & 48.21 & 50.78 & 44.58 \\
        BERT
        & 44.65 & 48.78 & 40.39  \\
        BERT-CNN
        & 50.31 & 50.60 & 49.72 \\
        \midrule
        
        BERT-CNN (full)
        & \textbf{53.69} & \textbf{54.42} & \textbf{53.38} \\
		\bottomrule	\end{tabular}}}
	\vspace{-0.5em}

	\caption{
	Comparison results 
	%of the baselines and our model 
	on the ITR dataset. 
	}
	\vspace{-1em}
	\label{tab:ITR_results}
	
\end{table}
\begin{table}[t]
	 \centering
{\renewcommand{\arraystretch}{0.8}
\resizebox{0.8\columnwidth}{!}
{
	\begin{tabular}[b]{
	%l c c c c c c
	lccc
	}
		\toprule
		\textbf{Methods}      & \textbf{Pre}  & \textbf{Rec} & \textbf{F1} \\
		\midrule

		MMSD
		& 76.57 & 84.15  & 80.18 \\
		%GRN (
         D\&R Net
        & 77.97 &  83.42 & 80.60 \\
        Res-BERT 
        & 78.87 & 84.46 & 81.57  \\
        Att-BERT 
        & 80.87 & 85.08 & 82.92  \\
         CMGCN
        & 83.63 & 84.69 & 84.16 \\
        MMSD-BERT 
        & 83.57 & 84.52 & 84.04 \\
        \midrule
        
        MMSD-BERT (full)
        & \textbf{85.50} & \textbf{85.92} & \textbf{85.70} \\
		\bottomrule	\end{tabular}}}
	\vspace{-0.5em}
	\caption{
	Comparison results 
	%of the baselines and our model 
	on the MSD dataset. 
	}
	\label{tab:MSD_results}
\end{table}
\begin{table}[t]
	 \centering
{\renewcommand{\arraystretch}{0.8}
\resizebox{0.7\columnwidth}{!}
{
	\begin{tabular}[b]{
	%l c c c c c c
	lccc
	}
		\toprule
		\textbf{Methods}      & \textbf{Pre}  & \textbf{Rec} & \textbf{F1} \\
		\midrule

		Xception
		& 56.0 & 54.5 & 54.4  \\
		%GRN (
        LSTM
        & 70.7 & 73.7 & 71.9  \\
        RoBERTa
        & 75.9 & 76.5 & 75.4  \\
        MMBT
        & 76.3 & 78.5 & 77.1 \\
        \midrule
        
        MMBT (full)
        & \textbf{79.15} & \textbf{79.88} & \textbf{78.76} \\
		\bottomrule	\end{tabular}}}
	\vspace{-0.5em}
	\caption{
	Comparison results 
	%of the baselines 
	%and our model 
	on the MHP dataset.\protect\footnotemark}
	\label{tab:MHP_results}
	\vspace{-1em}
\end{table}

%\paragraph{Baselines for MHP}
\paragraph{MHP Comparisons.}
Following the setup in  \citet{DBLP:conf/acl/BotelhoHV21}, we consider the Xception \cite{DBLP:conf/cvpr/Xception} baseline using visual features only.
%, which only encodes image content for classification, 
For the text-only comparison, 
%is used as the unimodal-image model while 
LSTM and RoBERTa \cite{DBLP:journals/corr/RoBERTa} are adopted.
%are utilized as the context-based unimodal-text models. 
MMBT \cite{DBLP:conf/acl/BotelhoHV21} is the SOTA model learning cross-modal representations with pre-trained MultiModal BiTransformers and will be employed as the base to experiment with our module.

%our base model, integrate the image features with text tokens to the 
%MultiModal BiTransformers, initialized with pre-trained
%BERT weights, for classification.

\footnotetext{
The baseline results are copied from the original paper, where the numbers are rounded to one decimal place.
%The results of baselines only have three significant figures, and the results of the full model also employ the same format for consistency.
}

\paragraph{Integrating our Comment-aware Self-training.}
Based on aforementioned SOTA architectures ( \textbf{base classifiers} --- CoMN-BERT, BERT-CNN, MMSD-BERT, and MMBT, selected for the four benchmarks), we further employ comment-aware self-training in their training and
%integrate our comment-aware self-training module with the 
%base classifiers,
%, which yield 
%The newly combined models are 
therefore result in CoMN-BERT (full), BERT-CNN (full), MMSD-BERT (full), and MMBT (full). 
%The model CoMN-BERT (full), BERT-CNN (full), NMSD-BERT (full), and MMBT (full) represent the corresponding base model adding the comment-aware self-training for MVSA, ITR, MSD, and MHP, separately. 

%To demonstrate the effects of different modules in the proposed retrieval-based comment-aware self-attention method, 
To further examine the relative contributions of each sub-module in comment-aware self-training, the following ablations are considered in comparison:
%we conduct experiments on four variants based on the base model:
(1) Base+Com,
integrating  BERT-encoded comment features in the base classifiers.
%utilize the retrieved comments to add context with the base model (Base+Com); 
(2) Base+ST,  self-training with retrieved tweets yet without comments. 
%with retrieved similar image-text pairs based on the base models
(3) Base+Com+ST, the full model without randomly dropping the retrieved comments in student model training.
%The full model
%use retrieved comments and self-training framework (Base+Com+ST);
%(4) randomly drop the retrieved comments for the student model in the self-training framework (Base+Com+ST+Drop).

\section{Experimental Discussions}

\subsection{Main Comparison Results}\label{ssec:exp:main}

\begin{table*}[t]
	 \centering
{\renewcommand{\arraystretch}{1.0}
\resizebox{2.0\columnwidth}{!}
{
	\begin{tabular}[b]{
	%l c c c c c c
	l |cc |c cc |ccc |ccc
	}
		\hline
		\multirow{2}*{\textbf{Model}}        & \multicolumn{2}{|c}{MVSA} &  \multicolumn{3}{|c}{ITR} & \multicolumn{3}{|c}{MSD} & \multicolumn{3}{|c}{MHP}\\
		
        & Acc & F1   &  Pre & Rec & F1
		& Pre & Rec & F1 & Pre & Rec & F1 \\
		\hline
        Base Classifier
        & 71.33 & 70.66 & 50.31 & 50.60 & 49.72 & 83.57 & 84.52 & 84.04 & 76.30 & 78.50 & 77.10 \\
        Base+Com
        & 72.34 & 71.57     & 52.08  & 52.67   & 51.64  & 84.76 & 85.19 & 84.98  & 77.31 & 78.29 & 77.67  \\
        Base+ST
        & 73.33 & 71.63     & 51.26 & 51.89 & 50.64   & 84.32 & 85.45 & 84.88 & 77.72 & 78.49 & 77.85\\
        Base+Com+ST
        & 73.11 & 72.29    & 53.06 & 53.45 & 52.32   &85.42 & 85.24 & 85.33  & 78.45 & 78.20 & 78.29  \\

        %Base+Com+ST+Drop (full)
        Full Model
        & 73.71 & 72.83  & 53.69 & 54.42 & 53.38 & 85.50 & 85.92 & 85.70 & 79.15 & 79.88 & 78.76 \\
		\hline	\end{tabular}}}
	\vspace{-0.5em}
	\caption{
	Ablation results on the four datasets. Our Full Model outperform all the ablations measured by all metrics.
	}
	\vspace{-1em}
	\label{tab:ablation_results}
\end{table*}

Table \ref{tab:MVSA_results}$\sim$\ref{tab:MHP_results} shows the main comparison results on MVSA, ITR, MSD, and MHP, respectively. 
%The observations are summarized as follows: 
%(1) The 

%We first observe 
The full model significantly outperforms
%improve the performance compared with 
all baselines and advances their base ablation on all test benchmarks (measured by paired t-test; $p-value<0.05$). 
This indicates that our
%the retrieved comments and 
comment-aware self-training can universally benefit varying tasks  
and classification architectures.
%retrieval-based comment-aware self-training method 
%can be employed on different models and different multimodal social media tasks. 
%Specially, our method could obtain gains 
It enables performance gains on both the simple architecture (e.g., BERT-CNN for ITR) and other more complicated models. 
%(i.e., CoMN-BERT for MVSA, NMSD-BERT for MSD, and MMBT for MHP). 
The possible reasons are twofold. 
%for the improvement 
First, retrieved comments, carrying viewpoints from human readers, may provide
%could provide 
complementary context hinting the cross-modal semantic understanding for weakly-connected image-text pairs.
%for the image-text pairs 
Second, our self-training may enable the models to leverage both labeled data and unlabeled retrieved data, potentially mitigating the overfitting issue caused by insufficient labeled data scales.
%more data are included for training by using the self-training framework; 

We also observe the models with BERT encoders 
%BERT-based models 
consistently outperform their counterparts with LSTM encoders, either in 
%LSTM-based models 
%on both 
multimodal or unimodal architectures. 
These demonstrate the benefit of pre-training on large-scale text, where the gained generic language understanding capability may enable models to well induce cross-modal meanings.
%enable promising results on downstream visual-lingual social media tasks.
%This demonstrates that pre-trained language models have the excellent ability to capture the semantics of texts.  

\subsection{Ablation study}\label{ssec:exp:ablation}
The general superiority of our method has been demonstrated in $\S$\ref{ssec:exp:main} compared to previous benchmark results.
Here we conduct an ablation study to further probe the relative contribution of varying components and 
%To examine the effects of different components of the proposed comment-aware self-training method, we conduct an ablation study and 
show the results 
%for the four tasks 
in Table \ref{tab:ablation_results}. 
%Both comments and self-training module contribute greatly to the model. 

The obvious performance drop of Base+Com and Base+ST, compared to the Full Model, together suggest the positive effects individually from retrieved comments and self-training.
These strengthen our previous findings: the comments may enrich context with human hints to bridge visual and lingual semantics and self-training may enrich the data scales with semantically related posts and comments to allow better robustness.  
%This demonstrates that comments could provide necessary context and bridge the images and texts while utilizing the self-training module to add semantically similar data into the training set is helpful for the four tasks. 

For the results of Base+Com+ST, though better than other ablations, are slightly worse than the Full Model.
It implies the extra benefit of modeling retrieved comments in self-training, while randomly dropping some of them may enable the student model to better catch up with the teacher, mitigating the least favorable perturbation phenomena \cite{DBLP:conf/cvpr/XieLHL20} in teacher-student alignment.

%Furthermore, from the results of Base+Com+ST and Base+Com+ST, adopting the strategy of randomly dropping comments could further boost the performance. 
%The reason might be that the student model with fewer comments would have more difficulty to align with the predictions of teacher model, which leads to favorable perturbation for the training step.

\subsection{Quantitative Analysis}

$\S$\ref{ssec:exp:ablation} shows the crucial roles self-training and comment retrieval play in our method. 
Here we further quantify the effects of varying unlabeled data scales on self-training and those of individual modality (images or text) on comment retrieval.

\begin{figure}[t]
\centering
\includegraphics[width=0.8\columnwidth]{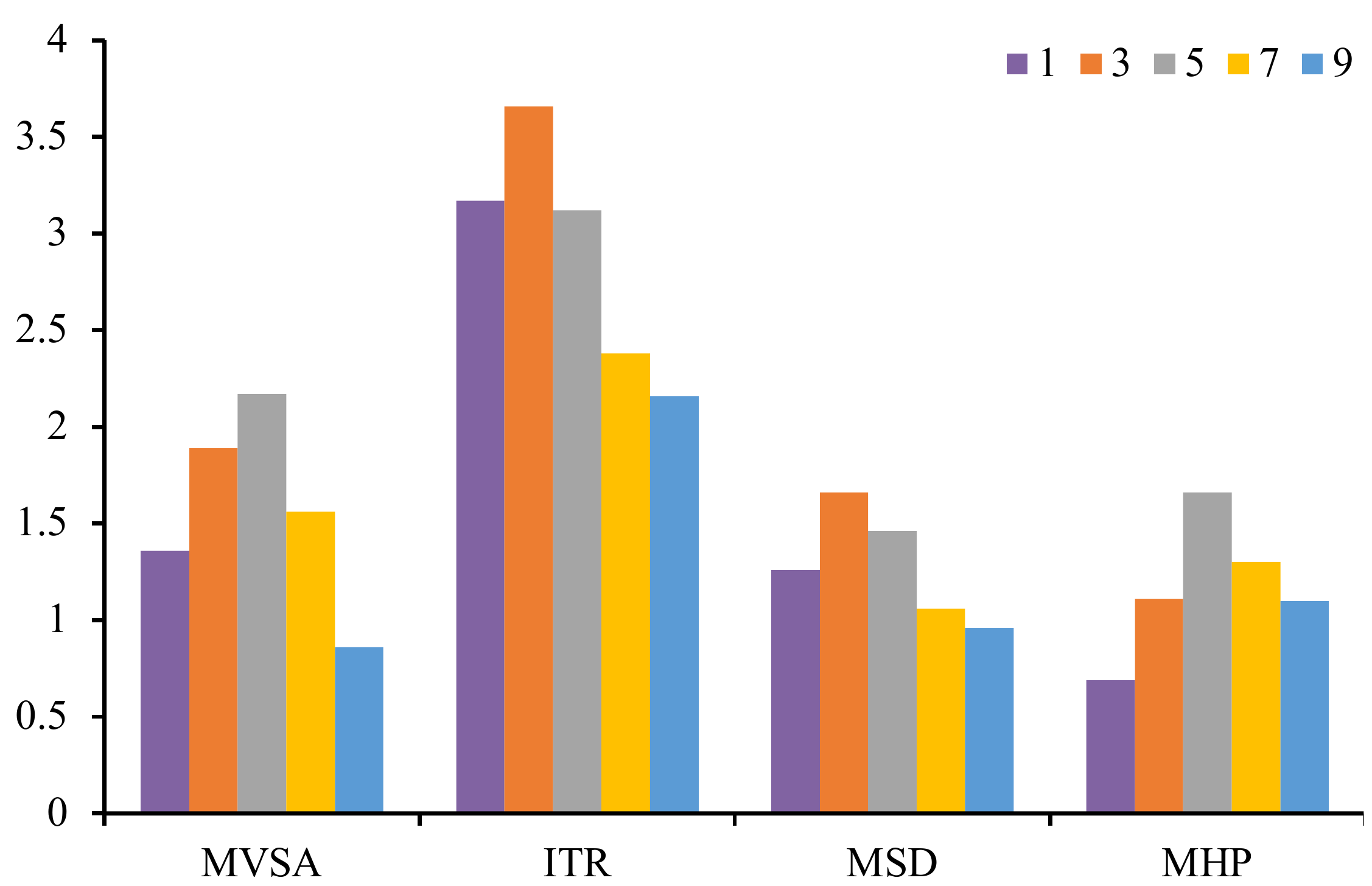}
\vspace{-0.5em}
\caption{Performance gain observed from self-training given varying number of unlabeled retrieved posts (and their associated comments).
%the retrieved image-text pairs used in the self-training. 
X-axis: within each dataset, the bars from left to right indicate self-training with varying number of posts ($K$); y-axis: the difference in F1 between our Full Model and Base Classifier.
%the different value of $K$. 
%Y-axis represents the performance gain compared with related base models.
}
\label{fig:self_train_num}
\end{figure}
\begin{figure}[t]
\centering
\includegraphics[width=0.8\columnwidth]{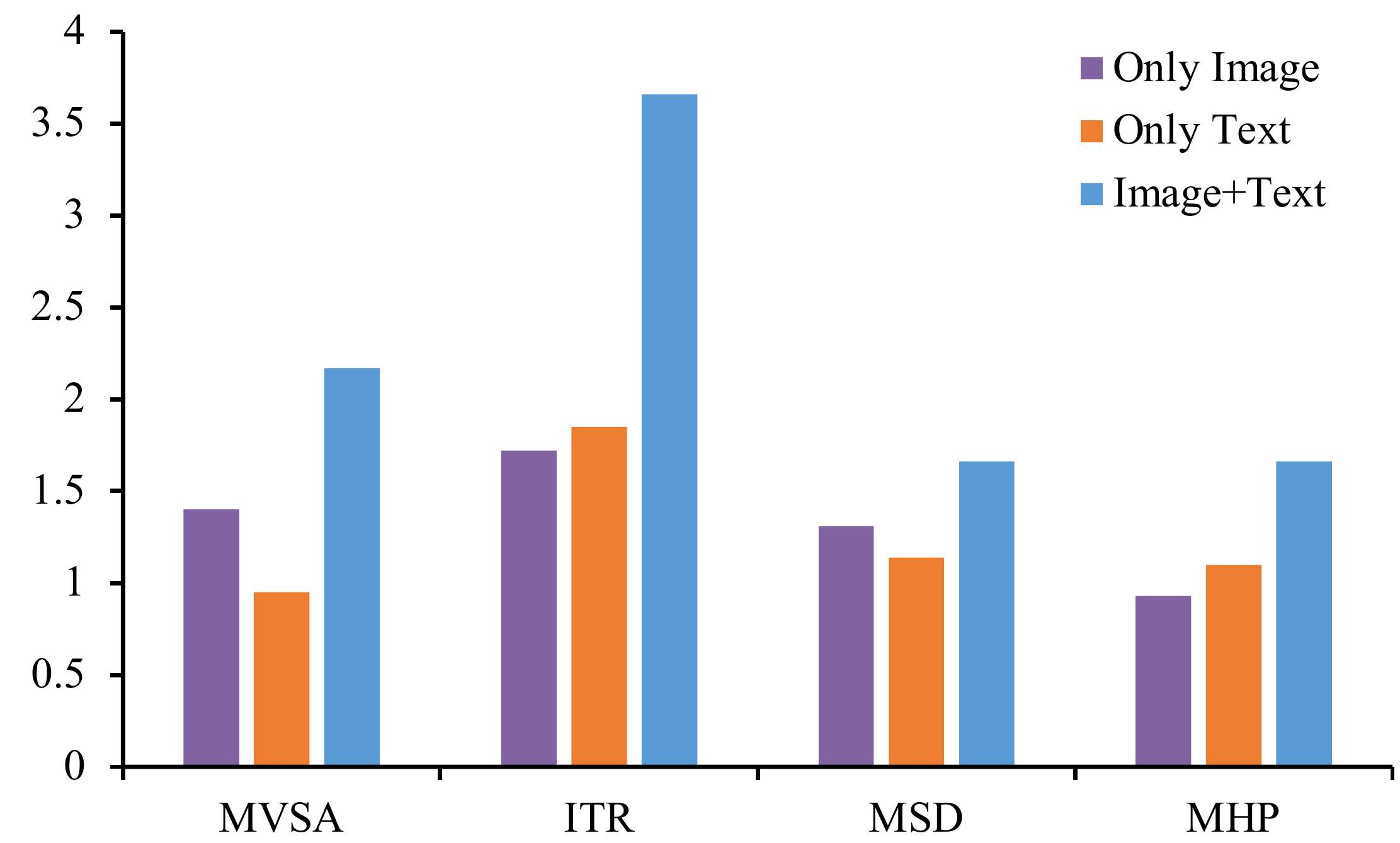}
\vspace{-0.5em}
\caption{F1 gain compared to the Base Classifier (y-axis) over varying datasets. For each dataset, the bars from left to right indicate the retrieval with image only, text only, and both image and text (Image+Text).  
%Full model performance compared with varying modality ablations for the four tasks. 
}
\vspace{-1em}
\label{fig:modality}
\end{figure}

%\paragraph{The effects of the num of retrieved image-text pairs used in the self-training}
\paragraph{Self-training w/ Varying Unlabeled Data Scales.}
%To examine model sensitivities to varying num of image-text pairs used for the step of self-training, 
Here we train our full model via self-training with varying number of retrieved posts ($K$) and show the performance gain compared to Base Classifier in Figure \ref{fig:self_train_num} (F1 difference of the Full Model and Base Classifier).
%setting the num $k$ of similar posts to $1,3,5,7,9$. 
%As shown in Figure \ref{fig:self_train_num}, 
%We observe that most models obtain the best performance when $k$ is 3 or 5. And the performance world decrease with more data (i.e., $k$=7 or 9). 
We observe the results peak at $K=3$ or $5$, implying  self-training may benefit from some similar unlabeled data while further retrieving more data may result in noise as well.

%The reason might be that there are not enough semantically similar posts in the retrieval dataset, and it's might be hard for the model to learn from the pseudo-labeled posts which are not similar to the original labeled data.

%\paragraph{The efftects of the modality used for the retrieval}
\paragraph{Modality Effects on Comment Retrieval.}
Recall that in comment retrieval, we balance the visual and lingual similarity to retrieve similar posts (and obtain their comments). 
To further study how features in varying modalities affect comment retrieval results, we examine two ablations relying on the similarity in image (Only Image) and text (Only Text) in comparison to the full model trading-off image and text similarities (Image+Text).

%To investigate the effectiveness of the proposed comment-aware self-training method when used with different retrieval modalities, we conduct experiments with only using image features, only using text features, and using image and text features for the retrieval process. 
The results (F1 gain compared to Base Classifier) are shown in Figure \ref{fig:modality}.
Varying tasks might prefer the similarity measure with image or text semantics,
whereas the full model leveraging posts' visual and lingual features achieves the best results.

%We can observe that the model with both modalities could consistently obtain the best performance on all tasks compared with the unimodality. This demonstrates that semantic gaps exists in the image-text pairs, and it's necessary to consider the both modalities for the understanding.

\subsection{Qualitative Analysis}\label{ssec:exp:qualitative}

The discussions above are from a quantitative view. 
To provide more insight, a case study will be presented here, followed by analyses for error cases.

% \begin{figure}[t]
% \centering
% \includegraphics[width=1.0\columnwidth]{figures/case_study.pdf}
% \caption{Visualization of attention heatmaps over the retrieved comments. From up to down shows the sampled cases from MVSA, ITR, MSD, and MHP benchmark. Deeper colors indicate higher attention weights. 
% }
% \label{fig:case_study}
% \end{figure}

\begin{figure}[t]
\centering
\includegraphics[width=0.9\columnwidth]{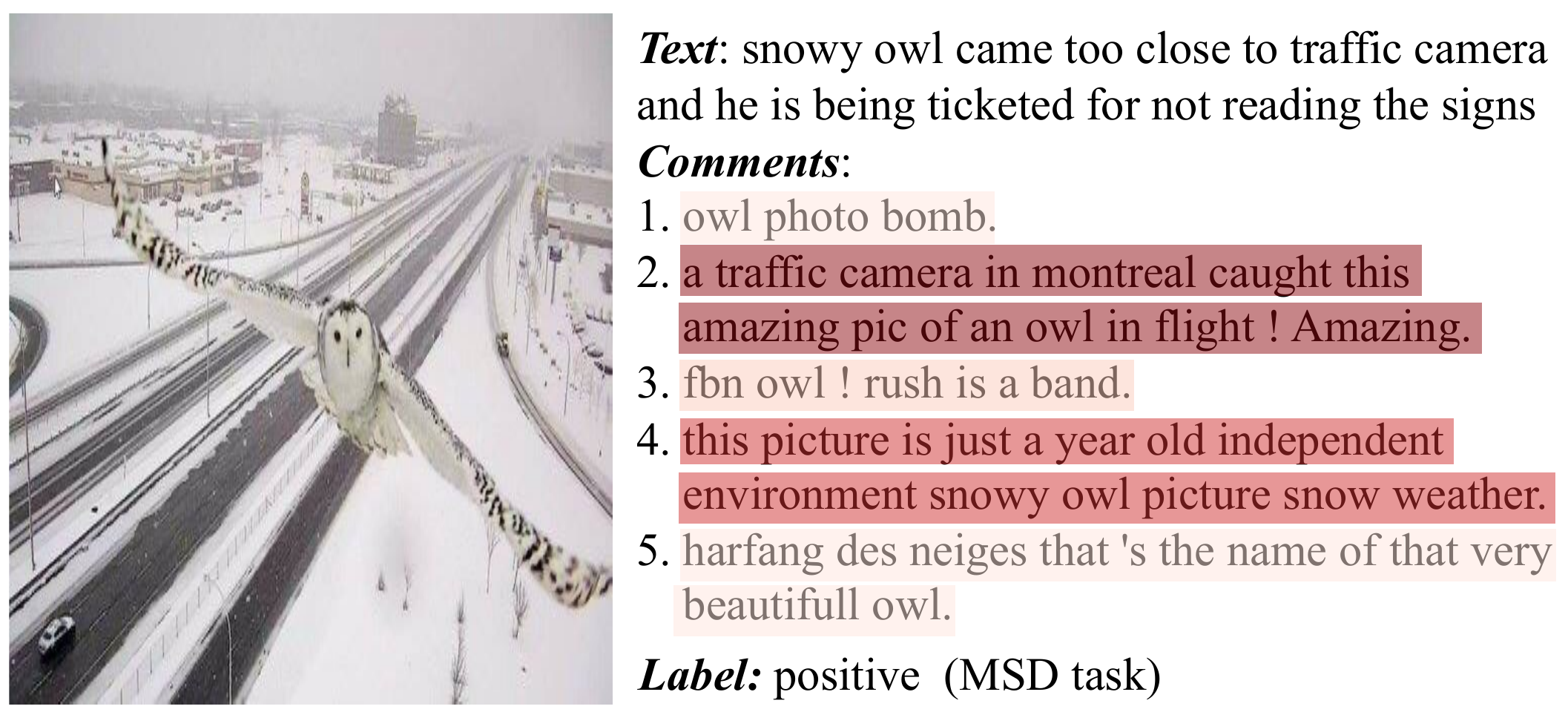}
\vspace{-0.5em}
\caption{Visualization of attention heatmaps over the retrieved comments for the MSD benchmark. Deeper colors indicate higher attention weights. 
}
% \vspace{-1em}
\label{fig:case_study}
\end{figure}

\paragraph{Case Study.}
% To analyze how comments hint cross-modal understanding,  attention maps over comments  are shown in Figure \ref{fig:case_study}, where a case is sampled from each benchmark. 
To analyze how comments hint at cross-modal understanding,  attention maps over comments  are shown in Figure \ref{fig:case_study}, where the case is sampled from the MSD benchmark. 
% \footnote{More cases could be found in Appendix \ref{appendix:case}.}
As can be seen, models tend to capture the salient comments mentioning the key visual objects, e.g., ``owl'' and ``traffic camera'' in the case, helpfully connecting visual semantics to lingual.
It is probably because human readers are likely to echo crucial points in their comments in response to what they viewed from a post, which inspires models to explicitize the weakly-connected visual-lingual semantics.

%We can observe that the retrieved comments could provide the necessary context from human views to fill the semantic gap between images and texts. Taking the third case for example, the text ``snowy owl is being ticketed for not reading the signs'', the model might be fused to distinguish the label of the post. However, the comment ``a traffic camera caught this amazing pic of an owl in flight'' could add external background context of the post. Therefore, the model can reasoning from the comments and  predict the true label.

\begin{figure}[t]
\centering
\includegraphics[width=0.9\columnwidth]{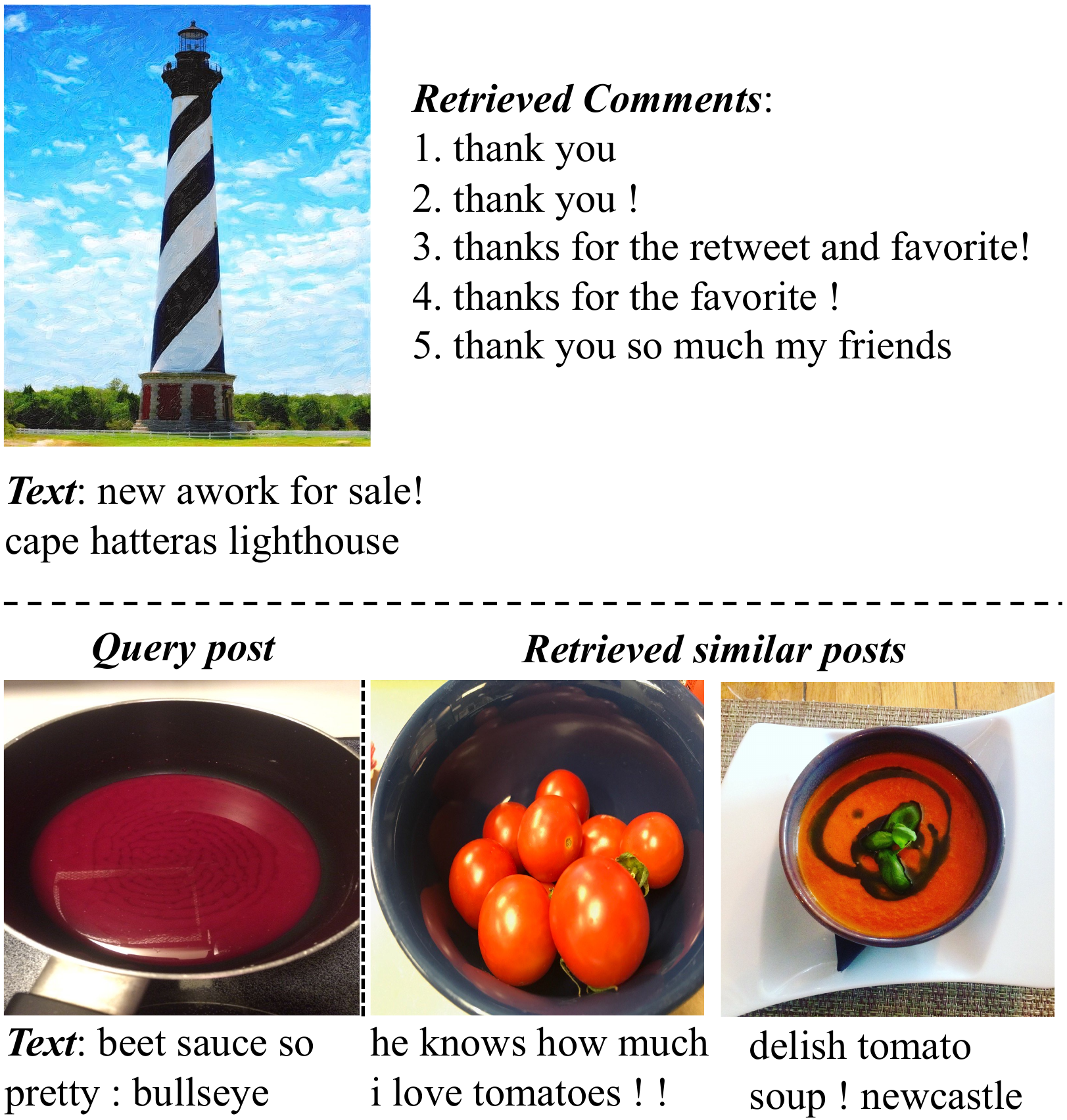}
\vspace{-0.5em}
\caption{Examples of major error types from comment and post retrieval. 
The top indicates the general comments and the bottom semantically unrelated posts. 
}
\vspace{-1em}
\label{fig:error_analysis}
\end{figure}

\paragraph{Error Analysis.}
%\input{figures/error_analysis}
%As mentioned above, the main gain of the proposed comment-aware self-training method come from the utilization of retrieval results. 
%Here we analyze error cases, which are mostly caused by 
The potential benefit has been potentially observed in varying cross-modal learning scenarios; however, many errors are also related to the retrieved comments and posts used in self-training.
Figure \ref{fig:error_analysis} summarizes the two major error types.
%Therefore, the quality of retrieval results could greatly influence the performance. 
%Here we summarize two main types of bad retrieval results which 
%are shown in : 

First, general comments, e.g., ``thank you'', are retrieved, useless in learning specific meanings in social media posts. This calls for a future direction for comment selection in a more effective manner.
%is general and useless for understanding the post. 
%the potential solution is 
%to design special rules to filter the retrieved comments. 
%
Second, semantically unrelated posts might be retrieved due to the misunderstanding to the query and hence result in irrelevant comments. 
For example, the posts concerning ``tomato'' are wrongly retrieved because of its similar color to ``beet sause''.
Future work may consider the advance in similarity measurement of cross-modal posts and the detection of high-quality unlabeled data 
(e.g., selecting the 
%highest-confidence 
pseudo-labeled data by confidence) for self-training.
 
%for self-training.

%The retrieved image-text pairs used for self-training are not related to the query post. 
%This reason might be that the similar posts are not included in the retrieval dataset, and the image encoder couldn't capture accurate objects. 
%Expanding the scale of the retrieval dataset and using more advanced image encoder could improve the quality of retrieved similar posts. Additionally, selecting the highest-confidence pseudo-labeled data instead of all retrieved similar posts might solve the dissimilar problem and improve the performance, which could be tried in future.

\section{Conclusion}
We have presented the potential of employing comments to better form visual-lingual understanding on social media, where 27M tweets with comments are contributed for the related study.
A novel framework is proposed to retrieve comments from similar posts and explore comments' hinting capabilities via self-training. 
Experimental results on four social media benchmarks show the universal benefit of leveraging retrieved comments and conduct comment-aware self-training on various multi-modal classification tasks and architectures. 

%proposed a retrieval-based comment-aware self-training framework for social media multimodal classification, where the retrieved comments are utilized to bridge the semantic gap between the images and texts while the retrieved semantically similar posts are employed in the self-training framework to solve the limitation of the scale of labeled data and improve the performance. Additionally, about 27M multimodal social meida tweets with comments are collected. Extensive experimental results on four multimodal social media tasks have shown the effectiveness of the proposed method.
\section*{Acknowledgements}
This paper is substantially supported by NSFC Young Scientists Fund (No.62006203), a grant from the Research Grants Council of the Hong Kong Special Administrative Region, China (Project No. PolyU/25200821), PolyU internal funds (1-BE2W, 4-ZZKM, and 1-ZVRH), and CCF-Baidu Open Fund (No. 2021PP15002000).

\section*{Limitations}
Here we point out more limitations in addition to what we have discussed in $\S$\ref{ssec:exp:qualitative}.

First, the post and comment retrieval should be built upon a large-scale corpus (wild dataset).
Substantial efforts might be needed to gather a likewise corpus, if applying our work to other a different social media platform. 
However, for our follow-up work exploring Twitter as well, the pre-trained retrieval module, based on the Faiss library, can work in an efficient manner.
%Although the retrieval efficiency is acceptable. 
For example, we test the retrieval module to search similar posts for 5,000 cross-media posts by using the Faiss library on one single 2080Ti GPU, and it would cost 231.66s for image modality and 77.68s for text modality.\footnote{The reason of different retrieval time is due to the different dimension of extracted features (i.e., the dimension is 2048 for image feature while 768 for text)} %Therefore, the retrieval efficiency is acceptable benefit from the Faiss library.

%The main limitation is the 
Second, the time and size of the retrieval corpus would result in another limitation.  
%constructed retrieval dataset. 
As shown in Table \ref{tab:retrieval_dataset_analysis}, we build the dataset with multimodal tweets 
%we collect the multimodal tweet data 
posted from 2014 to 2019. 
While a timely update might be needed if the task requires fresher data, e.g., the research of COVID-19 because the event becomes trendy in 2020.
%This means that no similar data might be retrieved from the retrieval dataset if the time of query data are not included in 2014 and 2019. 
%For example, it's impossible to retrieve the multimodal data about the COVID-19 due to the time limitation. 
%Additionally, not enough similar data can be retrieved for some unique multimodal tweets. 
%To solve the problem, the direct and simple way is to expand the size of the retrieval dataset and update the dataset with recent multimodal tweet data.
Nevertheless, dynamic dataset update might also explosively scale up the data quantity, and how to enable feasible real-time dataset management calls for another research question, which is beyond the scope of this paper and is valuable to be explored in future studies.

% Another limitation is the retrieval efficiency. We test the retrieval module to search similar posts for 5,000 cross-media posts by using the Faiss library on one single 2080Ti GPU, and it would cost 231.66s for image modality and 77.68s for text modality \footnote{The reason of different retrieval time is due to the different dimension of extracted features (i.e., the dimension is 2048 for image feature while 768 for text)}. Therefore, the retrieval efficiency is acceptable benefit from the Faiss library.
\section*{Ethical Considerations}
% We declare our dataset will cause no ethics problem.
% First, all retrieval data were collected in a
% manner which is consistent with the terms of use of standard data acquisition process regularized by Twitter API. The data is downloaded for the purpose of academic research.
% First, we follow the standard data acquisition process regularized by Twitter API. We downloaded the data for a purpose of academic research and is consistent with the Twitter terms of use. 

The benchmark datasets we experiment for classification are publicly  available with previous work, where the ethical concerns have been addressed by the authors of these original papers.

Our paper contributes a large-scale Twitter corpus (wild dataset) for post and comment retrieval.
%In this section, we focus on the main ethical considerations of the large-scale wide dataset.
We collected the data therein
%therein were collected 
%in a
%manner which is consistent with the 
following the terms of use of standard data acquisition process regularized by Twitter API. 
The data is downloaded only for the purpose of academic research.
Following the Twitter policy for datasets open-access, only the tweet IDs will be released.
Data requestors will be asked to sign a declaration form before accessing the data, 
making sure that the dataset will only be reused for research, under Twitter's policy compliance, 
%, as well as to declare that they will 
and not for collecting anything possibly raising ethical concerns, such as the sensitive and personal information. 

For our experiments, we have pre-processed to anonymize the data for privacy concern, e.g., removing authors' names and changing @mention and URL links to generic tags.
\bibliography{anthology}
\bibliographystyle{acl_natbib}
\clearpage

\end{document}